\documentclass[pdflatex,sn-mathphys-num]{sn-jnl}

\usepackage{graphicx}%
\usepackage{multirow}%
\usepackage{amsmath,amssymb,amsfonts}%
\usepackage{amsthm}%
\usepackage{mathrsfs}%
\usepackage[title]{appendix}%
\usepackage{xcolor}%
\usepackage{textcomp}%
\usepackage{manyfoot}%
\usepackage{booktabs}%
\usepackage{algorithm}%
\usepackage{algorithmicx}%
\usepackage{algpseudocode}%
\usepackage{listings}%
\usepackage{booktabs}
\usepackage{comment}
\usepackage{xspace}
\usepackage{xcolor}

\usepackage{float}
\usepackage{graphicx}
\usepackage{booktabs}

\theoremstyle{thmstyleone}%
\theoremstyle{thmstyletwo}%

\theoremstyle{thmstylethree}%

\usepackage{lineno}

\begin{document}

\newcommand{\methodname}{GUIDE-US\xspace}

\title[Grade-Informed Unpaired Distillation of Encoder Knowledge from Histopathology to Micro-UltraSound]{\methodname: Grade-Informed Unpaired Distillation of Encoder Knowledge from Histopathology to Micro-UltraSound}


\author*[2]{\fnm{Emma} \sur{Willis}}\email{emma.willis@queensu.ca}
\equalcont{These authors contributed equally to this work.}

\author[1]{\fnm{Tarek} \sur{Elghareb}}
\equalcont{These authors contributed equally to this work.}

\author[2]{\fnm{Paul F. R.} \sur{Wilson}}

\author[1]{\fnm{Minh Nguyen Nhat} \sur{To}}

\author[1]{\fnm{Mohammad Mahdi} \sur{Abootorabi}}

\author[2]{\fnm{Amoon} \sur{Jamzad}}

\author[3]{\fnm{Brian} \sur{Wodlinger}}

\author[2]{\fnm{Parvin} \sur{Mousavi}}

\author[1]{\fnm{Purang} \sur{Abolmaesumi}}

\affil[1]{\orgname{University of British Columbia}, \city{Vancouver}, \country{Canada}\footnote{P. Mousavi and P. Abolmaesumi are joint senior authors.}}

\affil[2]{\orgname{Queen's University}, \city{Kingston}, \country{Canada}}

\affil[3]{\orgname{Exact Imaging}, \orgaddress{\city{Markham}, \country{Canada}}}


\abstract{
\textbf{Purpose:} Non-invasive grading of prostate cancer (PCa) from micro-ultrasound (micro-US) could expedite triage and guide biopsies toward the most aggressive regions, yet current models struggle to infer tissue micro-structure at coarse imaging resolutions.

\textbf{Methods:} We introduce an unpaired histopathology knowledge-distillation strategy that trains a micro-US encoder to emulate the embedding distribution of a pretrained histopathology foundation model, conditioned on International Society of Urological Pathology (ISUP) grades. Training requires no patient-level pairing or image registration, and histopathology inputs are not used at inference.

\textbf{Results:} Compared to the current state of the art, our approach increases sensitivity to clinically significant PCa (csPCa) at 60\% specificity by 3.5\% and improves overall sensitivity at 60\% specificity by 1.2\%.

\textbf{Conclusion:} By enabling earlier and more dependable cancer risk stratification solely from imaging, our method advances clinical feasibility. Source code will be publicly released upon publication.
}




\maketitle


\section{Introduction}\label{sec1}
Prostate cancer (PCa) remains a leading cause of cancer-related death in men~\cite{GLOBOCAN2022}. Standard diagnosis involves ultrasound-guided prostate biopsy followed by histopathological examination of tissue samples. Accurate biopsy requires identification of suspected lesions in imaging where clinically significant PCa (sPCa) can be easily missed resulting in delayed treatment,  while indolent lesions may be difficult to distinguish from aggressive disease, leading to over-diagnosis. Aggressive PCa  must be detected early and treated for improved patient outcomes, while low grade disease can be monitored through active surveillance. 

Conventional ultrasound imaging has low sensitivity for PCa detection~\cite{ahmed2017diagnostic}. While multiparametric MRI (mpMRI) has higher sensitivity for cancer detection~\cite{ahmed2017diagnostic}, the requirement for dedicated facilities and expensive MRI machinery presents a major practical barrier to its application in many clinical settings. Micro-ultrasound (micro-US), which captures high resolution images of the prostate, has emerged as a promising, cost-effective and convenient alternative to MRI. A recent multi-center clinical trial showed that micro-US is non-inferior to mpMRI for csPCa detection~\cite{kinnaird2025microultrasonography}. Micro-US images require extensive expertise to interpret and are still limited by their low specificity for cancer detection. Hence, decision support tools to improve the interpretation of micro-US for PCa detection are needed.

Radiomics and deep learning methods have shown promise for PCa detection and risk stratification~\cite{Calace2022_Med,elghareb2025self}. The majority of these approaches use supervision from discrete labels (e.g., cancer versus benign, low-grade versus high-grade disease) originating from histopathological grade categories of corresponding biopsies. However, in reality, PCa is a complex and heterogeneous disease characterized by a spectrum of subtle changes in cellular structure and tissue microarchitecture. Discrete labels may result in models that fail to capture this nuance, reducing their ability to distinguish between healthy, clinically insignificant, and clinically significant disease. 
By comparison, in digital pathology, deep learning methods processing whole-slide image (WSI) have proven capable of learning rich feature representations that capture complex cellular and micro-architectural tissue properties, leading to accurate disease detection and grading~\cite{Bulten2022_NatMed_PANDA}. We therefore hypothesize that histopathology models can be used as a teacher to encourage micro-US models to learn more nuanced tissue characteristics to improve their downstream grading capabilities.

\emph{Multi-modal knowledge distillation} is a learning framework designed to build predictive models by leveraging complementary features across different data modalities. It has been extensively studied in the context of aligning natural images with language representations~\cite{liu2023clip}. Importantly, in such cross-modal settings the objective is typically to transfer \emph{semantic structure} (task-relevant organization of representations) rather than enforce pixel- or appearance-level similarity across modalities. In medical imaging, this framework can be adapted such that a pathology encoder transfers its knowledge to an imaging encoder, enabling the latter to capture histologic cues from imaging data alone during inference. With patient-paired cohorts, teacher--student feature matching has been used to transfer pathology features into MRI for classification (Deep Multimodal Guidance~\cite{mallya2022deep}). Correlation-based alignment methods learns a shared space between MRI and paired whole-mount histology, enabling the transfer of histologic signatures across modalities (CorrSigNIA~\cite{Bhattacharya2022CorrSigNIA}). Generative supervision approaches reconstruct or predict histology from radiographic latent representations, providing a rich target for distillation (HistoXGAN~\cite{Howard2024HistoXGAN}). Even in the absence of pixel-level registration, unaligned yet patient-paired synthesis can guide imaging encoders using ``histopathology-style'' features (ArtHiFy~\cite{Bhattacharya2024ArtHiFy}). However, these techniques rely heavily on \emph{paired} data, in the form of co-registered images from the same patient across different modalities. This requirement limits their direct application to PCa detection. First, datasets with micro-US and corresponding histopathology WSIs are not available. Second, there is an inherent modality gap both in spatial resolution (WSIs capture cellular detail, whereas micro-US resolves tissue microstructure) and the anatomical coverage (WSIs represent biopsy specimens, while micro-US encompasses the entire prostate).These gaps make direct structural matching across modalities ill-posed; instead, a practical goal is to transfer disease-relevant semantics (e.g., aggressiveness) that are shared across modalities. To effectively learn from histopathology, new methods must be developed that operate on more loosely paired data.

We propose \methodname, a novel method to train micro-US cancer assessment models with unpaired histopathology guidance. In this approach, we distill the knowledge of a pretrained WSI histopathology teacher model into a student micro-US image encoder at the distribution level. Our objective is not to align modality-specific appearance, but to encourage the micro-US embedding space to preserve \emph{grade-conditioned semantic structure} learned by the pathology teacher, using ISUP as a shared clinical anchor across modalities. To learn from unpaired data, we use a \emph{weak pairing} method, using a triplet loss to pull micro-US and histopathology representations of the same ISUP grade closer while pushing apart representations from different grades. To handle the gap in resolution and anatomical coverage between modalities, we utilize an attention-based multiple instance learning (ABMIL) approach to automatically learn the cross-modal correspondences before alignment. We integrate \methodname into a standard supervised learning pipeline by using a multi-task loss objective, demonstrating that the histopathology teacher significantly improves downstream PCa and csPCa detection. This design makes unpaired supervision meaningful because knowledge distillation is driven by shared grade semantics rather than structural correspondence. Our key contributions are as follows:

\begin{enumerate}
    \item We propose the first method, to the best of our knowledge, using image-histopathology knowledge distillation for micro-US PCa detection.
    \item We leverage ABMIL to address the scale and size disparity in image-histopathology knowledge distillation. This allows the model to prioritize regions in imaging that are most relevant to histopathology features. 
    \item Using a large, multi-center clinical trial dataset of micro-US, we demonstrate that \methodname achieves state-of-the-art PCa detection performance. 
\end{enumerate}

\section{Materials and Methods}\label{sec2}

\subsection{Data}

\paragraph{Micro-US Dataset} We utilized a private micro-US dataset comprising 578 patients from five clinical sites, collected prospectively under an approved trial (ClinicalTrials.gov: NCT02079025) with informed consent. Biopsies were performed on the ExactVu micro-US platform (Exact Imaging, Markham, Canada)~\cite{rohrbach2018high}. Each frame was linked to the corresponding histopathology of a biopsy core, including ISUP and percentage of cancer involvement. Most patients contributed 10--12 systematic cores from standard sextant locations.

In total, 6,607 cores were available; the ISUP distribution is summarized in Table~\ref{tab:micro-us-isup-distribution}. Binary masks delineating the biopsy needle tract and the prostate gland were also provided.

To avoid patient-level leakage, we performed \textbf{label-stratified} 5-fold cross-validation with splits defined at the patient level, ensuring that all cores from any individual appeared exclusively in either the training or validation set within a given fold. Stratification was performed on the patient-level labels to preserve a similar label distribution across folds, reducing the risk that a fold contains disproportionately few cases.

For preprocessing, B-mode images were resized via bilinear interpolation to \(1024 \times 1024\) pixels, normalized to \([0,1]\) on a per-image basis, and channel-repeated to form 3-channel inputs compatible with the foundation model.

\begin{table*}[t]
\centering
\caption{Micro-US ISUP grade-group distribution (number of cores).}
\label{tab:micro-us-isup-distribution}
\begin{tabular}{lrrrrrr}
\toprule
& Benign & ISUP 1 & ISUP 2 & ISUP 3 & ISUP 4 & ISUP 5 \\
\midrule
Total & 5727 & 0 & 480 & 195 & 134 & 71 \\
\bottomrule
\end{tabular}
\end{table*}

\paragraph{Histopathology Dataset}\label{subsubsec2} The histopathology dataset utilized in this study is derived from the Prostate Cancer Grade Assessment (PANDA) challenge~\cite{Bulten2022_NatMed_PANDA}, which includes 10,616 digital microscopic scanned slides of prostate core needle biopsies sourced from Radboud University Medical Center and Karolinska Institute. Each scanned slide is annotated with an ISUP grade indicating increasing levels of prostate cancer aggressiveness.

For quality control, we followed the preprocessing steps outlined in prior work~\cite{UNI2_FM}, excluding slides with erroneous annotations or noisy labels. This resulted in a refined dataset of 9,555 WSIs, with class distributions by ISUP grade and data split detailed in Table~\ref{tab:histopathology-distribution}. The dataset was label-stratified into 80:10:10 training-validation-test folds (7,647 training, 954 validation, and 954 test slides).

\begin{table*}[t]
\centering
\caption{PANDA ISUP distribution by split (number of WSIs).}
\label{tab:histopathology-distribution}
\begin{tabular}{lrrrrrr}
\toprule
 & Benign & ISUP 1 & ISUP 2 & ISUP 3 & ISUP 4 & ISUP 5 \\
\midrule
Train & 2083 & 1919 & 967 & 896 & 900 & 882 \\
Val   & 260  & 240  & 121 & 111 & 112 & 110 \\
Test  & 260  & 240  & 121 & 111 & 112 & 110 \\
\bottomrule
\end{tabular}
\end{table*}

\subsection{Methods}\label{sec3}
\begin{figure}[!t]
  \centering
  \includegraphics[width=\linewidth]{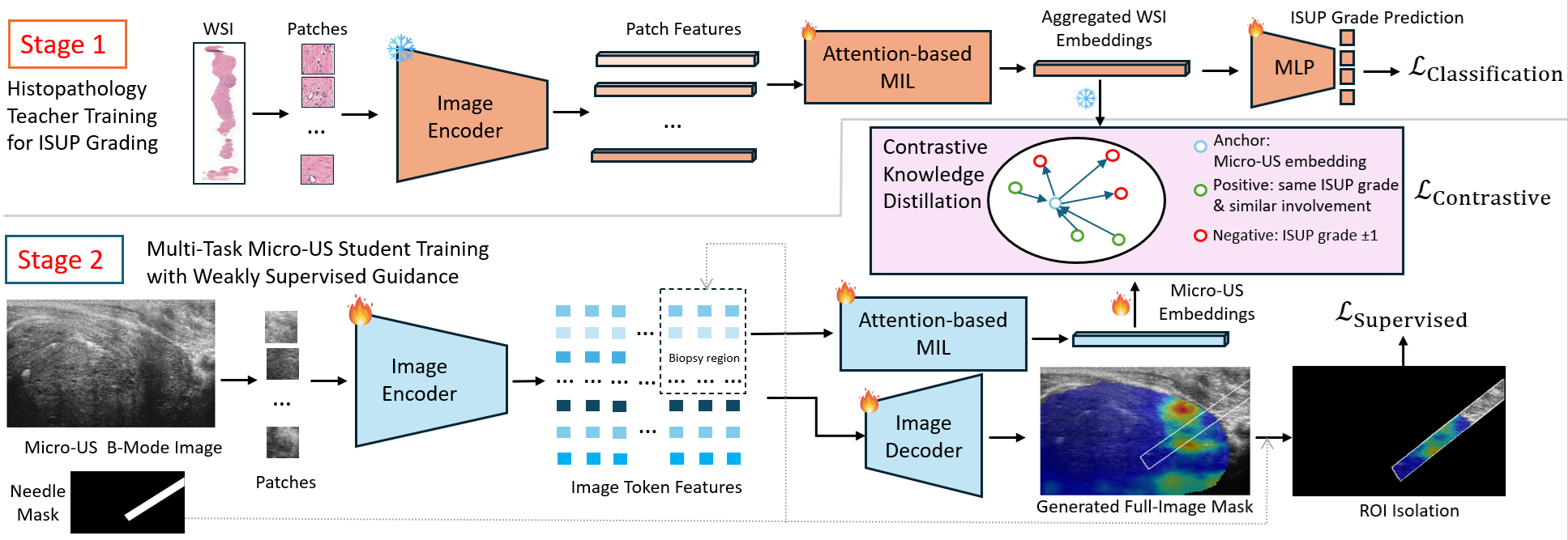}
\caption{Overview of the two-stage pathology-guided teacher-student framework for micro-US ISUP grading. In Stage 1 (top), a histopathology teacher model is trained on whole-slide images (WSIs) using attention-based multiple-instance learning (MIL) for aggregated patch embeddings and a multi-layer perceptron (MLP) for ISUP classification, with weights subsequently frozen. In Stage 2 (bottom), the micro-US student model undergoes multi-task training: an image encoder extracts the full image patches features, then the biopsy-region patches are aggregated via attention-based MIL into embeddings that are guided to match frozen pathology representations through contrastive knowledge distillation. Concurrently, an image decoder generates full-image cancer masks, enabling ROI isolation for weakly supervised segmentation guided by coarse histopathology reports.}
  \label{fig:architecture}
\end{figure}

\paragraph{Stage 1: Histopathology Teacher Model Training}\label{subsubsec:stage1}
An overview of our architecture is illustrated in Figure~\ref{fig:architecture}. We utilized publicly available slide-level patch embeddings generated by the GigaPath \cite{xu2024gigapath} computational pathology foundation model~\cite{UNI2_FM}, which was pre-trained on over 1.3 billion image tiles extracted from more than 170,000 proprietary histopathology slides. These embeddings were processed through the associated attention-based multiple-instance learning (AB-MIL) library~\cite{MIL_Transfer} to derive slide-level predictions for International Society of Urological Pathology (ISUP) grades.

\paragraph{Stage 2: Micro-US Student Model Training}\label{subsubsec:stage2}
In Stage 2, all the components of Stage 1 are frozen. We framed the micro-US grading as a multi-task learning problem, leveraging the pre-trained MedSAM model~\cite{medsam} within an encoder-decoder framework to perform weakly supervised segmentation guided by coarse histopathology reports, while simultaneously aligning representations with the frozen histopathology teacher through contrastive knowledge distillation. The segmentation branch restricts supervision to the biopsy footprint $\Omega_{\text{biopsy}}$, calculating loss exclusively within this region using the ground truth histopathology report as target, augmented by an involvement-aware loss function~\cite{wilson2025prostnfoundplus}. For pathology knowledge distillation, a micro-US image $\mathbf{I}\in\mathbb{R}^{1024\times1024}$ is tokenized with patch size $16$, yielding $64\times 64=4096$ non-overlapping tokens; indices $\mathcal{I}$ of tokens intersecting the needle-trace mask (typically $|\mathcal{I}|\approx 400$) form a bag $\mathcal{X}={\mathbf{x}t}{t\in\mathcal{I}}$, aggregated via an attention-based multiple-instance learning (AB-MIL) module to produce a single embedding $\mathbf{z}^{\text{us}}\in\mathbb{R}^D$. This $\mathbf{z}^{\text{us}}$ serves as the anchor in a contrastive triplet loss framework, with positive $\mathbf{z}^{\text{hist}}{+}$ sampled from histopathology embeddings matching the ISUP grade and involvement bin, and negative $\mathbf{z}^{\text{hist}}{-}$ as an adjacent-grade hard negative (ISUP differing by $\pm1$) to enhance grade separation. The triplet loss is optimized as
$$\mathcal{L}_{\text{triplet}} = \max \left( d(\mathbf{z}^{\text{us}},\mathbf{z}^{\text{hist}}_{+}) - d(\mathbf{z}^{\text{us}},\mathbf{z}^{\text{hist}}_{-}) + m, \, 0 \right),$$
where
$$d(\mathbf{u},\mathbf{v})=\lVert \mathbf{u}-\mathbf{v}\rVert_{2}$$
and margin $m=1.0$, with all embeddings $\ell_2$-normalized for metric stability. The overall objective combines these components:
$$\mathcal{L} = \mathcal{L}_{\text{seg}}(\Omega_{\text{biopsy}}) + \lambda\,\mathcal{L}_{\text{triplet}},$$
where $\lambda$ balances cross-modal alignment with weakly supervised segmentation.

\subsection{Experiments}
\paragraph{Baselines}
We compare against strong imaging baselines: (i) ProstNFound+ \cite{wilson2025prostnfoundplus} (current state-of-the-art micro--US classifier), (ii) SAM \cite{kirillov2023segment} with a task-specific head, (iii) MedSAM \cite{ma2024segment} fine-tuned on our data, and (iv) ProTeUS \cite{elghareb2025proteus}. We also include an \emph{imaging-only} variant of our architecture without any histopathology supervision (same encoder/decoder/ABMIL with only cross-entropy loss). Finally, we report results where distilled features from \methodname are fused with ProstNFound+.To assess robustness and statistical significance, we repeat training and evaluation with 8 independent random seeds under the same data split and protocol for both ProstNFound+ and \methodname.

\paragraph{Ablations}
We conduct two ablations to validate design choices. First, we swap the default triplet loss for a CLIP-style contrastive loss~\cite{radford2021learningtransferablevisualmodels}, which computes bidirectional cross-entropy between micro-US and histopathology embeddings at temperature $\tau=0.07$, averaging the losses for symmetry. Second, we evaluate robustness to the choice of pathology teacher encoder by swapping between UNI-2 \cite{UNI2_FM}, GigaPath \cite{xu2024gigapath}, and HistoSSLScaling \cite{Filiot2023ScalingSSLforHistoWithMIM} while keeping all other components fixed. Performance comparisons to the triplet baseline assess distillation efficacy across these variations.

\paragraph{Evaluation Metrics}
We report the following key metrics, averaged across 5-fold cross-validation: Area Under the Receiver Operating Characteristic Curve (AUROC) to evaluate overall discriminative performance; mean entropy within the biopsy region to quantify prediction confidence (lower values indicate more decisive outputs); sensitivity at 60\% specificity (Sens@60SPE) as a clinically relevant operating point balancing detection and false positives; and sensitivity at 60\% specificity for clinically significant prostate cancer (csPCa, $\text{ISUP} > 2$
; Sens@60SPE csPCa) to highlight efficacy on aggressive cases.

\paragraph{Implementation details}
To preserve the learned pre-trained weights and avoid catastrophic forgetting, only the adapter layers of the image encoder were fine-tuned. The image decoder and ABMIL model, configured with a projection dimension of 1024 and an attention dimension of 768, were trained end-to-end. We utilized the AdamW optimizer with a learning rate of $1 \times 10^{-4}$ and a batch size of 4. Training was conducted using five-fold cross-validation for up to 30 epochs per fold, with the optimal model selected based on balanced accuracy on the validation folds. Our approach was benchmarked against established methods, including ProstNFound+ \cite{wilson2025prostnfoundplus}, SAM \cite{kirillov2023segment}, MedSAM \cite{ma2024segment}, and ProTeUS \cite{elghareb2025proteus}. Additionally, we integrated our method with ProstNFound+ to evaluate the combined performance improvements and assess the overall efficacy of our framework.

\section{Results}\label{sec4}
\subsection{Quantitative Results}
\paragraph{Cancer Detection}

Table~\ref{tab:micro-US-results} summarizes the cancer detection quantitative performance metrics for the micro-US models based on the cancer heatmap output of the image decoder, including AUROC, mean entropy within the biopsy region, and sensitivity at multiple specificity operating points (Sens@40SPE, Sens@60SPE, and Sens@80SPE) for both prostate cancer (PCa) and clinically significant prostate cancer (csPCa, defined as ISUP\textgreater{2}). These metrics were averaged across the 5-fold cross-validation.

\begin{table*}[t]
\centering
\caption{Quantitative results for micro-US models. AUROC, Entropy, and Sensitivity at multiple specificity operating points are reported as the mean across 5-fold cross-validation.}
\label{tab:micro-US-results}
\resizebox{\textwidth}{!}{%
\begin{tabular}{lcccccccc} 
\toprule
& & & \multicolumn{2}{c}{\textbf{Sens@40\%Spec}} & \multicolumn{2}{c}{\textbf{Sens@60\%Spec}} & \multicolumn{2}{c}{\textbf{Sens@80\%Spec}} \\
\cmidrule(lr){4-5} \cmidrule(lr){6-7} \cmidrule(lr){8-9}
\textbf{Method} & \textbf{AUROC} & \textbf{Entropy} & \textbf{PCa} & \textbf{csPCa} & \textbf{PCa} & \textbf{csPCa} & \textbf{PCa} & \textbf{csPCa} \\
\midrule
SAM~\cite{kirillov2023segment} & 69.2 & 8.4 & 76.2 & 73.1 & 67.0 & 65.4 & 52.5 & 44.6 \\
MedSAM~\cite{kirillov2023segment} & 73.5 & 8.1 & 80.8 & 79.9 & 71.6 & 72.2 & 57.1 & 51.4 \\
ProTeUS~\cite{elghareb2025proteus} & 76.0 & 7.4 & 85.4 & 87.2 & 76.2 & 79.5 & 61.7 & 58.7 \\
ProstNFound+~\cite{wilson2025prostnfoundplus} & 77.9 & 7.32 & \textbf{89.4} & 92.1 & 79.1 & 82.5 & 64.5 & 60.2 \\
\textbf{Ours} & \textbf{78.4} & \textbf{6.78} & 89.3 & \textbf{92.6} & \textbf{80.3} & \textbf{86.0} & \textbf{64.7} & \textbf{64.3} \\
\bottomrule
\end{tabular}%
}
\end{table*}

Our proposed method, which integrates the histopathology distillation loss with the best-performing baseline model, ProstNFound+, demonstrates superior performance across all evaluated metrics. Specifically, it achieves an AUROC of 0.784, surpassing the state-of-the-art ProstNFound+. Notably, sensitivity at 60\% specificity for clinically significant cancer improves by approximately 3.5\%, which is critical in clinical settings as it facilitates earlier detection of aggressive disease while maintaining acceptable specificity, thereby reducing unnecessary biopsies and associated patient morbidity. Additionally, the mean entropy within the biopsy region decreases from 7.32 to 6.78, reflecting greater model confidence in predictions. Lower entropy signifies more decisive and localized probabilistic outputs, which is essential for interpretability in diagnostic workflows, enabling clinicians to focus on high-confidence regions and improving the reliability of AI-assisted triage in prostate cancer detection. We further report a Wilcoxon signed-rank test on Sens@60SPE csPCa across matched 8 random-seed runs, showing that the improvement is statistically significant ($p=0.0234$). These gains underscore the value of unpaired histopathology knowledge distillation in bridging scale and modality gaps, ultimately advancing toward more precise, non-invasive grading.

\begin{figure}[t]
\centering
\includegraphics[width=0.8\textwidth]{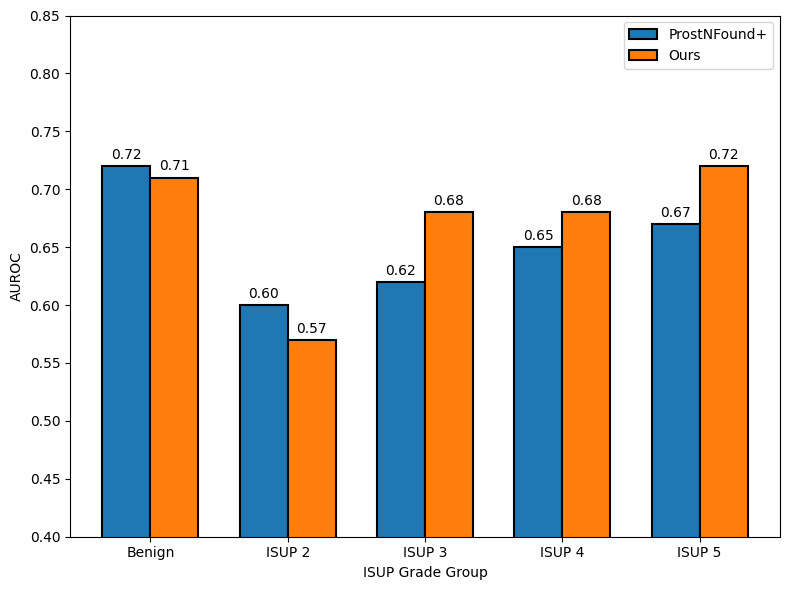}
\caption{AUROC per ISUP grade group for biopsy-region features from the encoder, comparing our method (orange) to the ProstNFound+ baseline (blue).}
\label{fig:microus-auroc-per-isup}
\end{figure}

\paragraph{Cancer Grading}
To evaluate the impact of histopathology distillation of learned representations on PCa grading, we assessed the classification accuracy of features extracted from the biopsy region using the micro-US image encoder, before and after knowledge distillation. Figure~\ref{fig:microus-auroc-per-isup} illustrates the AUROC per ISUP grade group for our method compared to the ProstNFound+ baseline.
As shown, our method achieves higher AUROC across all grades, with an average improvement of 4.5\% in clinically significant cancer types. These gains are particularly pronounced for life-threatening aggressive cancers (e.g., ISUP 3, 4, and 5), highlighting the clinical relevance of enhanced feature quality through distillation.

\subsection{Ablation Studies}

\begin{figure}[t]
  \centering
  \includegraphics[width=\linewidth]{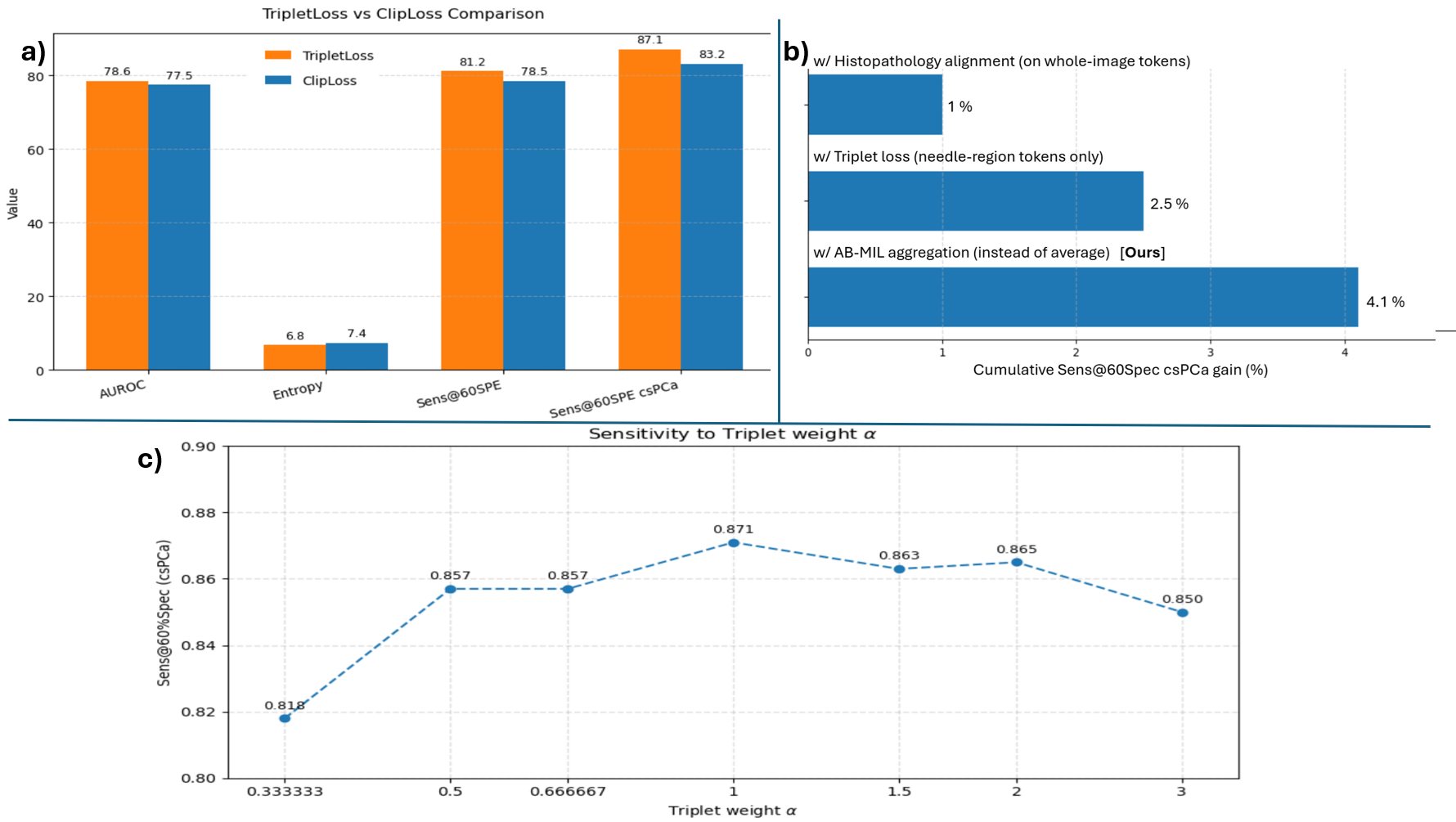}
  \caption{Ablation studies. (a) Triplet loss vs.\ CLIP-style contrastive loss across key metrics. (b) Step-wise ablation of proposed components showing cumulative gain in Sens@60\%Spec (csPCa) from the ProstNFound+ baseline. (c) Sensitivity to triplet-loss weight $\alpha$ (mean across folds) on Sens@60\%Spec (csPCa).}
  \label{fig:ablation_merged}
\end{figure}

\noindent\textbf{Distillation loss function.}
Figure~\ref{fig:ablation_merged}\,(a) presents a comparison of key performance metrics between the triplet loss and the CLIP-style contrastive loss in our knowledge distillation framework.
The triplet loss outperforms the CLIP loss across all evaluated metrics. This superior performance underscores the effectiveness of triplet loss in facilitating better knowledge distillation from unpaired histopathology to micro-US representations. Theoretically, since the datasets are inherently unpaired and positive samples are randomly selected based on matching ISUP grade and involvement, the CLIP loss---which treats all other batch samples as negatives---can introduce confusion when incidental samples in the batch share similar attributes, inadvertently penalizing the embeddings. In contrast, the triplet loss enables explicit selection of hard negatives from adjacent ISUP grades, ensuring focused separation of grade boundaries and more robust distillation without batch-induced artifacts, thereby enhancing the distillation process and yielding more discriminative features.

\vspace{4pt}
\noindent\textbf{Step-wise ablation of proposed components.}
Figure~\ref{fig:ablation_merged}\,(b) isolates the contribution of each proposed module, starting from the imaging-only baseline (ProstNFound+). We then progressively add: (i) histopathology alignment on \emph{full micro-US image tokens}, (ii) restricting triplet alignment to \emph{needle/biopsy-region tokens only}, and (iii) replacing average pooling with \textbf{AB-MIL} aggregation. The results show monotonic improvements in Sens@60\%Spec (csPCa) as each component is introduced.

\vspace{4pt}
\noindent\textbf{Sensitivity to triplet-weight $\alpha$.}
Figure~\ref{fig:ablation_merged}\,(c) reports sensitivity to the triplet-loss weight $\alpha$ (relative to the segmentation term), averaged across folds. Performance is stable across a broad range around $\alpha \in [0.5, 2]$, with the best result at $\alpha=1$, while more extreme settings reduce performance.

\vspace{4pt}
\noindent\textbf{Robustness to pathology teacher encoder.}
To test whether gains depend on a specific pathology foundation model, we repeat the full pipeline using the \emph{same histopathology tiles} while swapping the feature extractor. In addition to UNI-2 (main results), we evaluate GigaPath and HistoSSLScaling. Table~\ref{tab:teacher_robustness} shows comparable performance

\begin{table}[htbp]
\centering
\caption{Teacher robustness: swapping the pathology teacher encoder (same tiles, different feature extractors).}
\label{tab:teacher_robustness}
\small  
\begin{tabular}{lcccc}
\toprule
\textbf{Image Encoder} & \textbf{AUROC} & \textbf{Entropy} & \textbf{Sens@60\%Spec} & \textbf{Sens@60\%Spec (csPCa)} \\
\midrule
UNI-2 & 78.6 & 6.75 & 81.2 & 87.1 \\
GigaPath & 78.0 & 6.5 & 82.0 & 87.7 \\
HistoSSLScaling & 78.2 & 6.8 & 80.5 & 85.6 \\
\bottomrule
\end{tabular}
\end{table}

\subsection{Qualitative Results}
\begin{figure}[t]
\centering
\includegraphics[width=\textwidth]{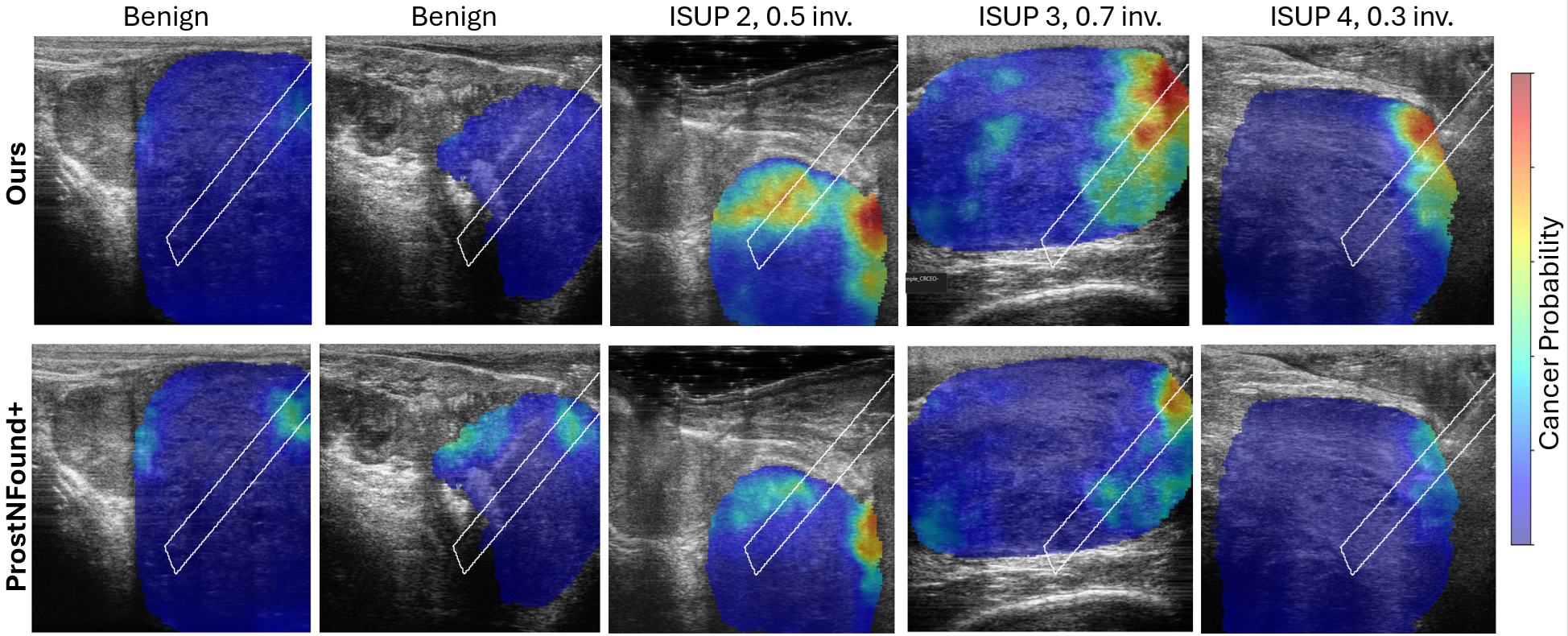}
\caption{Qualitative comparison of cancer probability heatmaps between our method (top row) and the ProstNFound+ baseline (bottom row) on micro-US cases. The biopsy regions are overlaid on each image. The ground-truth ISUP grade and cancer involvement for each case are shown.}
\label{fig:micro-US-qualitative}
\end{figure}

Figure~\ref{fig:micro-US-qualitative} presents a qualitative comparison of cancer probability heatmaps generated by our proposed method and the most competitive baseline, ProstNFound+, across representative micro-US cases, including benign and varying ISUP grades and cancer involvements. The heatmaps overlay predicted cancer probabilities on the original ultrasound images, with color intensity indicating confidence levels (blue for low probability, red for high).
In the benign cases (first two columns), our method produces markedly lower activations along the biopsy needle trace, minimizing erroneous high-probability regions that could lead to unnecessary interventions—a common challenge in baseline models. For the ISUP 2 case, our histopathology-enhanced approach yields more confident and precisely localized heatmaps, with elevated activations concentrated in suspicious areas, facilitating improved risk stratification for indolent disease. Notably, in the higher-grade ISUP 3 and ISUP 4 cases, the baseline model exhibits low to negligible detections, potentially missing clinically significant cancers, while our method robustly highlights malignant regions with high probability and more accurate cancer involvement. We hypothesize that this is the result of histopathology knowledge distillation enhancing sensitivity to subtle microstructural cues. These visualizations highlight the clinical utility of our framework in providing interpretable, reliable outputs that align more closely with histopathological ground truth.

\section{Conclusion}\label{sec5}

We introduced an unpaired pathology knowledge distillation approach that, paired with attention, teaches imaging encoders to focus on csPCa-relevant regions and features. In our experiments, this increased AUROC and raised sensitivity at matched specificity, suggesting practical gains for earlier, more reliable risk stratification and more precise biopsy targeting. These results validate the approach of using histopathology encoders as “teachers” even without paired datasets. This could enable histopathology-aware, potentially automatic cancer grading from imaging alone. Our method incurs no additional deployment cost since the histopathology branch is only required during training, not inference. In future work, we will test alternative knowledge distillation objectives and smarter negative-sample selection, and run multi-center prospective studies to evaluate robustness. If successful, this line of work could bring pathology-level insight into routine imaging workflows and help clinicians act sooner for patients most at risk.

\section*{Acknowledgments}
This work was supported by the Natural Sciences and Engineering Research Council of Canada (NSERC) and the Canadian Institutes of Health Research (CIHR). Parvin Mousavi is supported by Canada CIFAR AI Chair and the Vector Institute. We would also like to thank the computing resources provided by the Advanced Research Computing (ARC) at the University of British Columbia in supporting this project.

\section*{Disclosure of Interests}
Brian Wodlinger is Vice President of Clinical and Engineering at Exact Imaging and provided access to the dataset used in this study. No other author has any potential conflict of interest to disclose.

\bibliography{sn-bibliography}

\end{document}